# GPU-accelerated image alignment for object detection in industrial applications

Trung-Son Le and Chyi-Yeu Lin

Department of Mechanical Engineering
National Taiwan University of Science and Technology
43 Keelung Road, Section 4, Taipei, Taiwan
D10203810@mail.ntust.edu.tw, jerrylin@mail.ntust.edu.tw

*Abstract*—This research proposes a practical method for detecting featureless objects by using image alignment approach with a robust similarity measure in industrial applications. This similarity measure is robust against occlusion, illumination changes and background clutter. The performance of the proposed GPU (Graphics Processing Unit) accelerated algorithm is deemed successful in experiments of comparison between both CPU and GPU implementations

*Keywords— occlusion; illumination; alignment; texture-less; pose space search*

## I. Introduction

Pattern matching, also referred to as template matching, is a large branch of computer vision technologies. Its solutions are generally governed by the characteristics of the surface of working objects which are generally classified into with texture and texture-less. Texture is defined by how an object appearance looks and feels e.g. surface characteristics of the object. Such features are successfully described and represented by vision algorithms like SIFT [1], [2] and its variants [3] computed from visual sensor information. Invariant feature-based approach in object recognition has mature literature and results and is beyond the scope of this paper. Texture-less approach refers to shape-based object recognition which was pioneered by Ullman and his research group [4]–[6]. Ullman addresses that object recognition by its shape is the most dominant recognition approach in model-based object recognition [4].

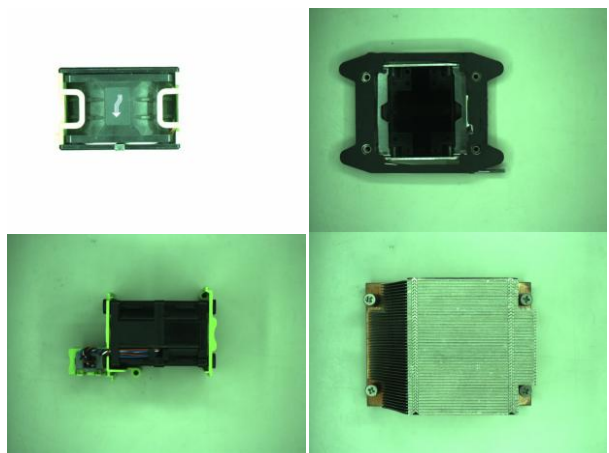
Fig. 1. Objects with few texture

In robotic assembling of industrial parts, the task commonly involves objects without sufficient surface characteristic e.g. objects with few textures or even no texture such as transparent objects. Mostly, these objects either have silhouettes/edges as significant features or are in extreme illumination condition or have repetitive features as shown in Fig. 1.

Besides, there are strict requirements on reliability of the systems; matching result should be performed consistently under a wide range of changing conditions of illumination, occlusions and types of surface textures. The solutions for textured objects have already reached a high level of robustness [2], [7], [8]. However, none of them has claimed similar performance on featureless objects

The alignment method can meet industrial assembling conditions and requirements due to simplicity and robustness. This alignment-based approach is expected to work well on low-texture parts. Industrial assembling also works with constrained environment i.e. object placements are on the planar surface and the straight-view from the camera can be guaranteed with small deviations.

Some of the first researches using alignment methods can date back to late 1950s with the applications on character recognition [9]. Followed by the 1980s, research groups, led by Shimon Ullman and Daniel Huttenlocher [5], [6], [10], and later Rucklidge [11]–[13], W. J. and Olson C. F. [14], carried out outstanding research and successfully published many great papers during two decades. Their approaches evolved from correspondence-based to non correspondence-based. They gradually tackled and solved the most difficult multi-view problems in object recognition from 3D, affine/weak perspective and full perspective. The scope of the objects which they can deal with upgraded from flat planar objects, to near-flat objects, and finally full 3D objects.

In recent years, alignment approach is revived by Steger group and Halcon [15]–[18] and then becomes one of the competing commercial packages in the industry. Lately, the approach which is integrated with probabilistic techniques has challenged to work on full 3D objects of translucent material under arbitrary view of Hinterstoisser, *et al* [19]–[22].

This paper employs the geometric based template matching using alignment approach which focuses on the planar object recognition. The implementation aims to attain a practical performance solution by using up-to-date graphics processing hardware power and builds a foundation for extensive research towards this direction. In following sections, this paper discusses and compares the algorithm performance in both CPU and in GPU implementations.

## II. METHODS

In this section, we will describe how the detection is programmed, beginning with a preprocessing of input data, followed by a target searched throughout whole pose space in both implementation versions.

The detection algorithm receives two images including a model to be detected and a search image; tuning parameters such as step size in each dimension x-translation, y-translation, theta-rotation; edge detection thresholds; number of pyramid levels; neighboring-search size. The smaller the step size, the finer the search grid is, however, it accordingly increases the search space and processing time. The edge detection thresholds ensure enough edge information is captured in model and search images regardless of lighting conditions. The number of pyramid levels should be sufficiently small such that the top-pyramid image still reserves enough details for correct detection. The currently used neighboring-search size is 3. The algorithm starts with an edge extraction of the input images. The generating x-gradients, y-gradients and gradient magnitudes of edge points are recorded. Edge extraction is followed by an image pyramid construction. Finally, the calculation of the detection score starts with the pyramid's top image.

With CPU implementation, our algorithm realizes an exhaustive 3D pose space search with 3 nested iterative loops for each dimension (x-translation, y-translation, and theta-rotation). At each pose, a score is calculated and is recorded in case it is a better score. This results in a maximum score and its corresponding pose after the whole space is searched through. To calculate a pose score, an iteration through all edge points is performed to accumulate a voting score for each edge point. During the algorithm tests, we face false detections in highly cluttered backgrounds. This can be explained by the difference between theoretical matching score and its digitized implementation in practice where search steps are not continuous and background clutter can be incorrectly detected as template features. To resolve this problem, a neighboring search nearby each edge point is added to find its maximum voting score. This results in a significant score increase when the search template is nearby to the target, and therefore, it can outscore the mistaken score of a highly cluttered background (see Fig. 3).

GPU's implementation takes advantage of its multi-core power. Our object detection has iteration spaces that can be split to thousands of GPU cores to speed up. Thrust library provides a straight-forward interface to implement nested parallelism which can be easily mapped to our algorithm. We use the first Thrust function transform_reduce to distribute pose space with each pose score calculation to GPU threads and return a maximized score. In its transform functor, the second transform_reduce distributes voting score calculation over all edge points and returns the similarity score given in equation (number). The third nested transform_reduce is inside the second and is used to maximize the edge-point's voting score over its neighboring. This nested hierarchy can be illustrated in Fig. 2.

Over the iteration spaces, a robust similarity measure $M$ is computed by a normalized score summation of the edge orientation consistency from each pair of projected $p_i$ and working image $w_i$ edge points out of $n$ total points.

$$M = \frac{1}{n}\sum_{i=1}^{n} \frac{p_i \cdot w_i}{\|p_i\| \cdot \|w_i\|} \quad (1)$$

This similarity measure uses each edge point as a feature. Despite some of these edge points are occluded, the rest points still provide a matching score. This offers the capability to deal with occlusion, illumination and clutter. The implementation is based on oriented edge pixel which also reduces significantly false alarm of matching.

There are possibilities of implementing different types of projection from template edge point $t_i$ to the projected edge point $p_i$. This paper applies a 2D rigid transformation $(u_x, u_y, \theta)$.

$$\begin{bmatrix} p_x \\ p_y \end{bmatrix} = \begin{bmatrix} \cos\theta & -\sin\theta \\ \sin\theta & \cos\theta \end{bmatrix} \begin{bmatrix} t_x \\ t_y \end{bmatrix} + \begin{bmatrix} u_x \\ u_y \end{bmatrix} \quad (2)$$

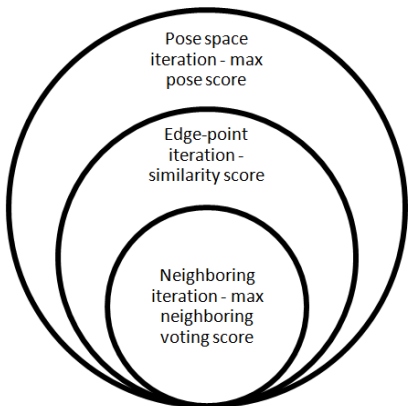

Fig. 2. Nested search space

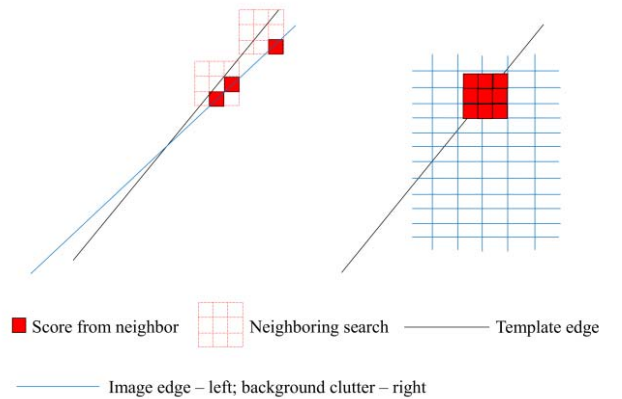

Fig. 3. Neighboring search to resolve clutter background false detection

## III. EXPERIMENTS

To evaluate the robustness of our similarity measure, the tests are performed and compared between CPU and GPU versions at different lighting conditions, occlusion and clutter. Our test set is manually created with 7 samples taken by Basler camera using a setup with straight-angle camera and four bar lights shown in Fig. 4. Each sample includes a pair of template and working images at 812-by-617 pixels in resolution. The algorithm runs on an ASUS laptop CPU Intel Core i3, 8GB RAM, NVIDIA GeForce 610M graphics unit running Ubuntu OS.

Fig. 5 shows detection results on 3 out of 7 collected samples. The detected template is overlaid on its working image. The same search range is used for all samples including 813-pixel x-translation, 618-pixel y-translation, 90-degree rotation with their search steps being 3 pixel, 3 pixel and 3 radian respectively. All these samples show correct detections regardless of illumination change, occlusion and clutter.

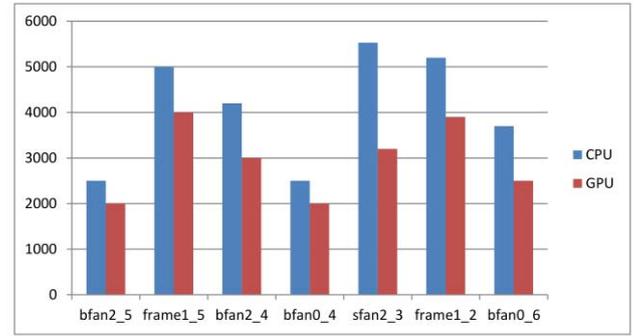

Fig. 6. Processing time (in milliseconds) comparison between CPU and GPU versions on 7 samples

To evaluate the improvements of GPU over CPU versions, Fig. 6 compares processing time of both implementations in 7 samples. In general, GPU version improves 20-30% computation time of its CPU counterpart.

## IV. CONCLUSION

This paper proposed a GPU-accelerated edge-based template matching using the alignment approach as a practical solution for industrial assembly implementation that can be applied robustly with low computing time. The algorithm is implemented in parallel processing with GPU to improve the running time which are tested and compared with CPU version. The test results have shown dominant GPU performance and robust detection capabilities under lighting changes, occlusion and background clutter. In future, this algorithm will be tested against a bigger dataset, with bigger workloads so as to demonstrate full potential of GPU advantages.

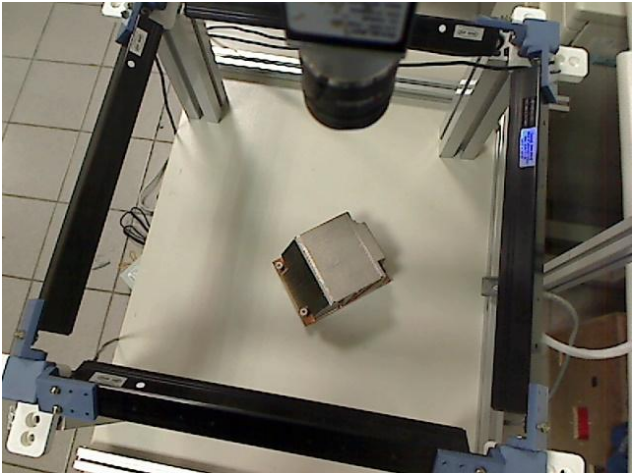

Fig. 4. Experiment setup

ACKNOWLEDGMENT

The work of this paper is supported by Ministry of Science and Technology, Taiwan (ROC) under the grant MOST 104-2221-E-011 -059 -MY3.

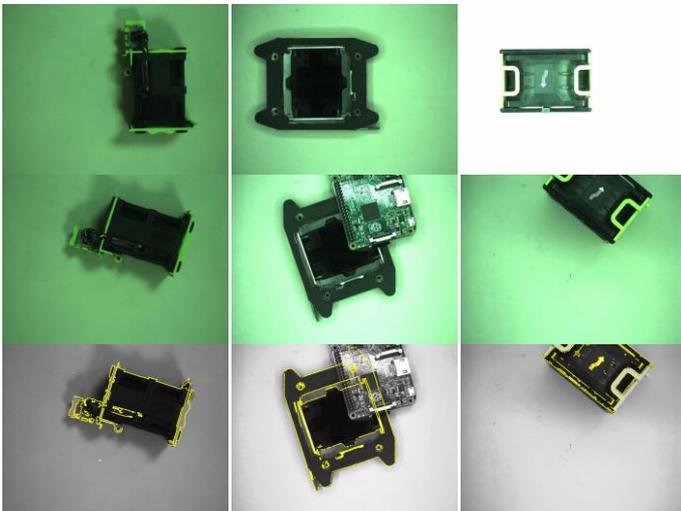

Fig. 5. Three out of 7 detection results; First row – templates; Second row – working images; Third row – detected templates are overlaid in working images; First column – sfan2_3; Second column – frame1_2; Third column – bfan2_5